\documentclass[11pt,a4paper]{article}
\usepackage{authblk}
\usepackage[hyperref]{acl2021}
\usepackage{times}
\usepackage{latexsym}
\usepackage{varwidth}
\usepackage{graphicx}
\usepackage{amsfonts}
\usepackage{subfigure}
\DeclareUnicodeCharacter{3000}{}

\usepackage{bm}
\usepackage{booktabs}
\usepackage{amsmath}
\usepackage{amssymb}
\usepackage{algorithm}
\usepackage{algorithmic}
\usepackage{adjustbox}
\usepackage{multirow}
\usepackage{multicol}
\usepackage{array}
\usepackage{booktabs}
\usepackage{pifont}
\usepackage{float}
\usepackage{microtype}

\aclfinalcopy 
\def\aclpaperid{3946} 

\usepackage{times,flushend}
\usepackage[normalem]{ulem}

\usepackage{etoolbox}
\makeatletter
\patchcmd{\maketitle}
{\def\@makefnmark}
{\def\@makefnmark{}\def\useless@macro}
{}{}
\makeatother

\title{Sketch and Refine: Towards Faithful and Informative Table-to-Text Generation}

\author[ ]{Peng Wang}
\author[ ]{Junyang Lin}
\author[ ]{An Yang}
\author[ ]{\\Chang Zhou}
\author[ ]{Yichang Zhang}
\author[ ]{Jingren Zhou}
\author[$$\dag]{Hongxia Yang\thanks{\llap{\textsuperscript{\dag}}Corresponding author.}}
\affil[ ]{DAMO Academy, Alibaba Group}
\affil[ ]{\tt \{zheluo.wp, junyang.ljy, ya235025, ericzhou.zc, }
\affil[ ]{\tt {yichang.zyc, jingren.zhou, yang.yhx \}@alibaba-inc.com}}

\date{}

\begin{document}
\maketitle
\begin{abstract}
Table-to-text generation refers to generating a descriptive text from a key-value table. 
Traditional autoregressive methods, though can generate text with high fluency, suffer from \textit{low coverage} and \textit{poor faithfulness} problems.
To mitigate these problems, we propose a novel \textbf{S}keleton-based two-stage method that combines both \textbf{A}utoregressive and \textbf{N}on-\textbf{A}utoregressive generation (\textbf{SANA}). 
Our approach includes: (1) skeleton generation with an autoregressive pointer network to select key tokens from the source table; (2) edit-based non-autoregressive generation model to produce texts via iterative insertion and deletion operations. 
By integrating hard constraints from the skeleton, the non-autoregressive model improves the generation's coverage over the source table and thus enhances its faithfulness. 
We conduct experiments on both the WikiPerson and WikiBio datasets.
Experimental results demonstrate that our method outperforms the previous state-of-the-art methods in both automatic and human evaluation, especially on coverage and faithfulness. 
In particular, we achieve PARENT-T recall of 99.47 in WikiPerson, improving over the existing best results by more than 10 points.
\end{abstract}

\section{Introduction}
\begin{figure}[t]
    \centering
    \includegraphics[width=1.0\linewidth]{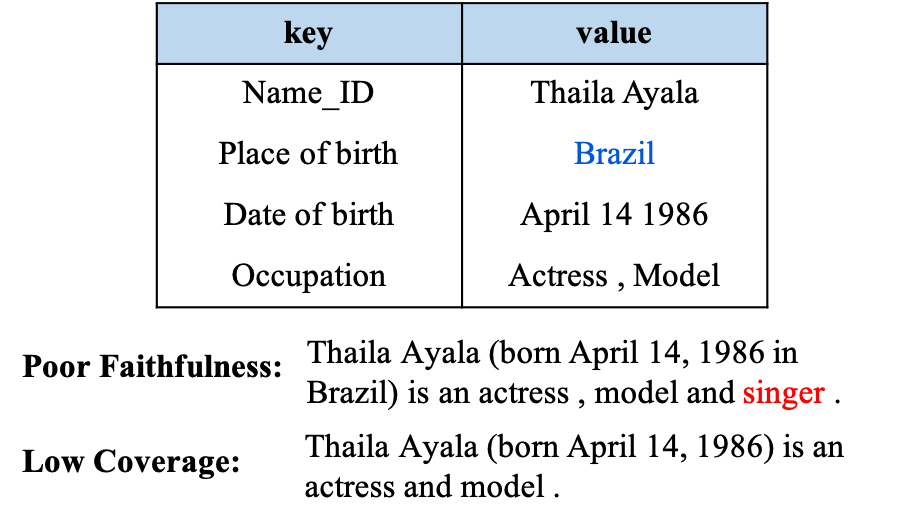}
    \caption{An example of table-to-text generation. The case of \textit{poor faithfulness} hallucinates content not entailed by the table (marked in red color). The case of \textit{low coverage} misses the information of the person's birth place (marked in blue color).}
    \label{fig:example1}
\end{figure}

Table-to-text generation is a challenging task which aims at generating a descriptive text from a key-value table.
There have been a broad range of applications in this field, such as the generation of weather forecast \cite{mei2016talk}, sports news \cite{wiseman2017challenges}, biography \cite{lebret2016neural,wang2018describing}, etc.
Figure \ref{fig:example1} illustrates a typical input and output example of this task.

Previous methods \cite{liu2018table,nie2018operation,bao2018table} are usually trained in an end-to-end fashion with the encoder-decoder architecture \cite{bahdanau2014neural}. 
Despite generating text with high fluency, their lack of control in the generation process leads to \textit{poor faithfulness} and \textit{low coverage}. 
As shown in Figure \ref{fig:example1}, the case of \textit{poor faithfulness} hallucinates the occupation ``singer" not entailed by the source table, and the case of \textit{low coverage} misses the information of the place of birth.
Even if trained with a cleaned dataset, end-to-end methods still encounter these problems as it is too complicated to learn the probability distribution under the table constraints \cite{parikh2020totto}. 


To alleviate these problems, recent studies \citep{shao2019long,puduppully2019data_content,ma2019key} propose two-stage methods to control the generation process.
In the first stage, a pointer network selects the salient key-value pairs from the table and arranges them to form a content-plan. 
In the second stage, an autoregressive seq2seq model generates text conditioned on the content-plan. 
However, such methods can cause the following problems:
(1) since the generated content-plan may contain errors, generating solely on the content-plan leads to inconsistencies;
(2) even if a perfect content-plan is provided, the autoregressive model used in the second stage is still prone to hallucinate unfaithful contents due to the well-known \textit{exposure bias} \cite{wang2020exposure} problem;
(3) there is no guarantee that the selected key-value pairs can be described in the generated text. 
As a result, these methods still struggle to generate faithful and informative text.

In this paper, we propose a \textbf{S}keleton-based model that combines both \textbf{A}utoregressive and \textbf{N}on-\textbf{A}utoregressive generation (\textbf{SANA}). 
SANA divides table-to-text generation into two stages: \textit{skeleton construction} and \textit{surface realization}.
At the stage of skeleton construction, an autoregressive pointer network selects tokens from the source table and composes them into a skeleton. We treat the skeleton as part of the final generated text.
At the stage of surface realization, an edit-based non-autoregressive model expands the skeleton to a complete text via insertion and deletion operations. 
Compared with the autoregressive model, the edit-based model has the following advantages: 
(1) the model generates text conditioned on both the skeleton and the source table to alleviate the impact of incomplete skeleton;
(2) the model accepts the skeleton as decoder input to strengthen the consistency between the source table and generated text;
(3) the model generates texts with the hard constraints from the skeleton to improve the generation coverage over the source table. 
Therefore, SANA is capable of generating faithful and informative text. 

The contributions of this work are as follows:
\begin{itemize}
    \item We propose a skeleton based model \textbf{SANA} which explicitly models skeleton construction and surface realization. 
    The separated stages helps the model better learn the correlation between the source table and reference.
    \item To make full use of the generated skeleton, we use a non-autoregressive model to generate text based on the skeleton. 
    To the best of our knowledge, we are the first to introduce non-autoregressive model to table-to-text generation task.
    \item We conduct experiments on WikiPerson and WikiBio datasets. Both automatic and human evaluations show that our method outperforms previous state-of-the-art methods, especially on faithfulness and coverage. Specially, we obtain a near-optimal PARENT-T recall of 99.47 in the WikiPerson dataset.
\end{itemize}

\section{Related Work}
\paragraph{Table-to-text Generation} Table-to-text generation has been widely studied for decades \cite{kukich1983design,goldberg1994using,reiter1997building}. 
Recent works that adopt end-to-end neural networks have achieve great success on this task \cite{mei2016talk,lebret2016neural,wiseman2017challenges,sha2018order,nema2018generating,liu2018table,liu2019hierarchical}. 
Despite generating fluent texts, these methods suffer from poor faithfulness and low coverage problems. 
Some works focus on generating faithful texts. 
For example, \citet{tian2019sticking} proposes a confident decoding technique that assigns a confidence score to each output token to control the decoding process.
\citet{filippova2020controlled} introduces a ``hallucination knob” to reduce the amount of hallucinations in the generated text.
However, these methods only focus on the faithfulness of the generated text, they struggle to cover most of the attributes in the source table.

Our work is inspired by the recently proposed two-stage method \cite{shao2019long,puduppully2019data_content,moryossef2019step,ma2019key,trisedya2020sentence}. 
They shows that table-to-text generation can benefit from separating the task into content planing and surface realization stages. 
Compared with these methods, SANA guarantee the information provided by the first stage can be preserved in the generated text, thus significantly improving the the coverage as well as the faithfulness of the generated text.

\paragraph{Non-autoregressive Generation}
Although autoregressive models have achieved remarkable success in natural language generation tasks, they are time-consuming and inflexible. 
To overcome these shortcomings, \citet{gu2017non} proposed the first non-autoregressive (NAR) model that can generate tokens simultaneously by discarding the generation history. 
However, since a source sequence may have different possible outputs, discarding the dependency of target tokens may cause the degradation in generation quality. This problem also known as the ``multi-modality'' problem \cite{gu2017non}. 
Recent NAR approaches alleviate this problem via partially parallel decoding \cite{stern2019insertion,sun2019fast} or iterative refinement \citep{lee2018deterministic,ghazvininejad2019mask,gu2019levenshtein}. 
Specially, \citet{stern2019insertion} performs partially parallel decoding through insertion operation.
\citet{gu2019levenshtein} further incorporates deletion operation to perform iterative refinement process.
These edit-based models not only close the gap with autoregressive models in translation task, but also makes generation flexible by allowing integrates with lexical constrains.
However, the multi-modality problem still exists, making it difficult to apply NAR models to other generation tasks, such as table-to-text generation, story generation, etc.
In this work, we use the skeleton as the initial input of our edit-based text generator. The skeleton can provide sufficient contexts to the text generator, thus significantly reducing the impact of multi-modality problem.

\section{Methods}
\begin{figure*}[t]
    \centering
    \includegraphics[width=1.0\linewidth]{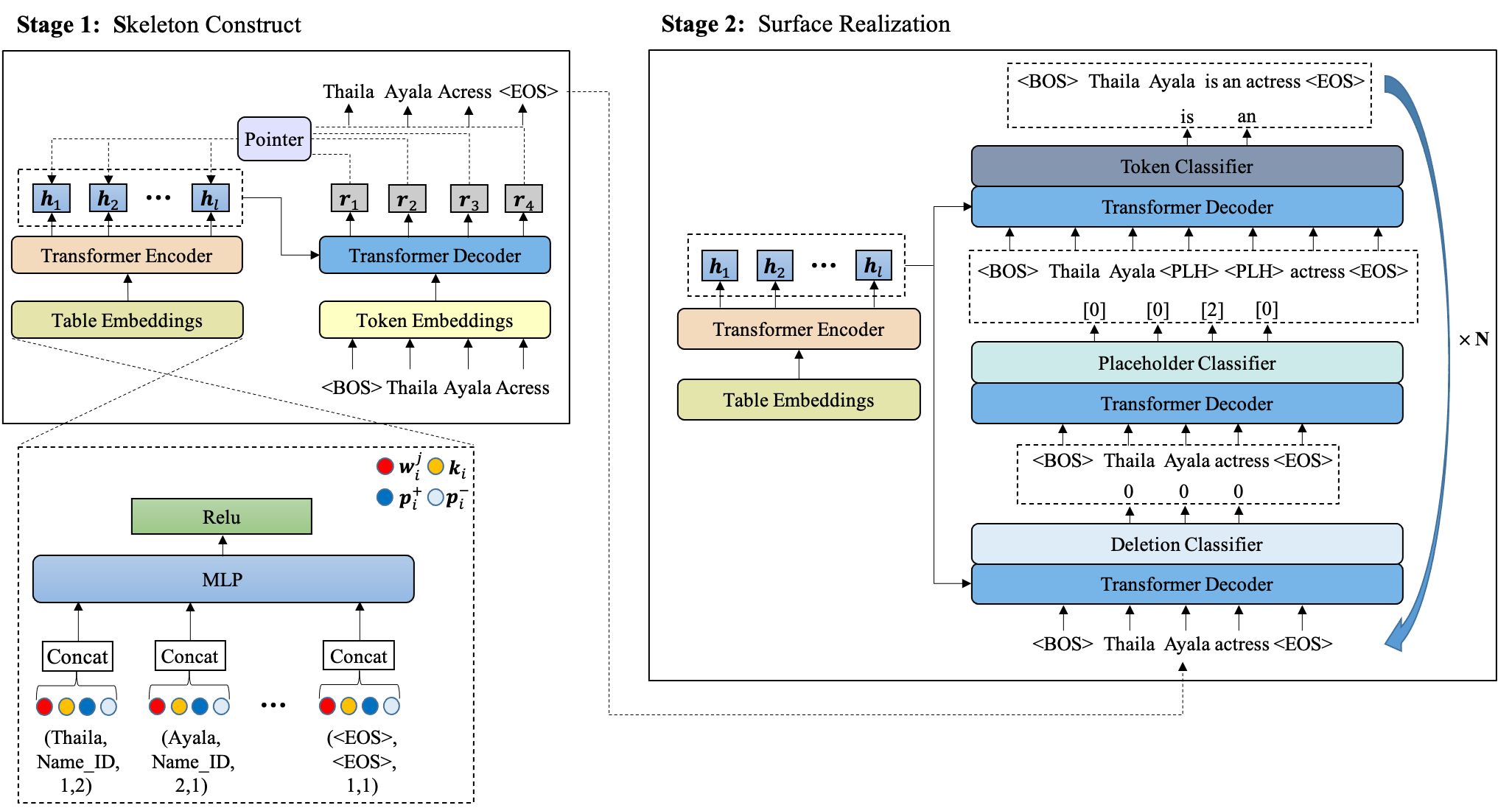}
    \caption{The overall diagram of SANA for generating description for \textit{Thaila Ayala} in Fig \ref{fig:example1}.}
    \label{fig:framework}
\end{figure*}

The task of table-to-text generation is to take a structured table $T$ as input, and outputs a descriptive text $Y = \{y_1, y_2, ..., y_n\}$. 
Here, the  table $T$ can be formulated as a set of attributes $T = \{a_1, a_2, ..., a_m\}$, where each attribute is a key-value pair $a_i = \langle k_i, v_i \rangle$.

Figure \ref{fig:framework} shows the overall framework of SANA. It contains two stages: skeleton construction and surface realization.
At the stage of skeleton construction, we propose a Transformer-based \cite{vaswani2017attention} pointer network to select tokens from the table and compose them into a skeleton. 
At the stage of surface realization, we use an edit-based Transformer to expand the skeleton to a complete text via iterative insertion and deletion operations. 

\subsection{Stage 1: Skeleton Construction}
\subsubsection{Table Encoder}
The source table is a set of attributes represented as key-value pairs $a_i = \langle k_i, v_i \rangle$. Here, the value of an attribute $a_i$ is flattened as a token sequence $v_i = \{w_i^1, w_i^2, ..., w_i^l\}$, where $w_i^j$ is the $j$-th token and $l$ is the length of $v_i$.
Following \citet{lebret2016neural}, we linearize the source table by representing each token $w_i^j$ as a 4-tuple $(w_i^j, k_i, p_i^+, p_i^-)$, where $p_i^+$ and $p_i^-$ are the positions of the token $w_i^j$ counted from the beginning and the end of the value $v_i$, respectively. For example, the attribute of ``$\langle \textit{Name\_ID, \{Thaila Ayala\}} \rangle$" is represented as ``(\textit{Thaila, Name\_ID}, 1, 2)" and ``(\textit{Ayala, Name\_ID}, 2, 1)". 
In order to make the pointer network capable of selecting the special token $\langle EOS \rangle$\footnote{$\langle EOS \rangle$ denotes the end of the skeleton.}, we add a special tuple $(\langle EOS \rangle, \langle EOS \rangle, 1, 1)$ at the end of the table.

To encode the source table, we first use a linear projection on the concatenation $\left[\bm{w}_i^j;\bm{k}_i;\bm{p}_i^+;\bm{p}_i^-\right]$ followed by an activation function:
\begin{equation}
\label{equ:Table representation}
\bm{f}_i^j = {\rm Relu}(\bm{W}_f[\bm{w}_i^j;\bm{k}_i;\bm{p_i^+};\bm{p_i^-}] + \bm{b}_f)
\end{equation}
where $\bm{W}_f$ and $\bm{b}_f$ are trainable parameters. Then we use the Transformer encoder to transform each $\bm{f}_i^j$ into a hidden vector and flatten the source table into a vector sequence $\bm{H}=\{\bm{h}_1, \bm{h}_2, ..., \bm{h}_l\}$.

\subsubsection{Pointer Network}
After encoding the source table, we use a pointer network to directly select tokens from the table and compose them into a skeleton. Our pointer network uses a standard Transformer decoder to represent the tokens selected at the previous steps. Let $\bm{r}_t$ denote the decoder hidden state of previous selected token $\hat{y}_t$. The pointer network predict the next token based on the attention scores, which are computed as follows:
\begin{align}
\alpha_{ti} &=  \frac{e^{u(\bm{r}_t,\bm{h}_i)}}{\sum_{j=1}^{l}e^{u(\bm{r}_t,\bm{h}_j)}} \\
u(\bm{r}_t,\bm{h}_i) &= \frac{(\bm{W}_q\bm{r}_t)\cdot(\bm{W}_k\bm{h}_i)}{\sqrt{d_{r}}}
\end{align}
where $\bm{W}_q$ and $\bm{W}_k$ are trainable parameters, $d_r$ is the embedding dimension of $\bm{r}_t$. According to the calculated probability distribution $\bm{\alpha}$, we select token based on the following formula:
\begin{align}
P_{copy}(w) &= \sum_{w=w_i} \alpha_{ti} \\
\hat{y}_{t+1} &= \mathop{\arg\max}_{w}P_{copy}(w)
\end{align}
where $\hat{y}_{t+1}$ represents the output at the next timestep, and $P_{copy}(w)$ represents the probability of copying token $w$ from the source. There may be multiple identical tokens in the table, so we sum up the attention scores of their corresponding positions.


The pointer network needs target skeletons as supervision, which are not provided by the table-to-text datasets. In this paper, we obtain the skeleton by collecting tokens in both the table and description except the stop words. The token order in the skeleton remains the same as their relative positions in the description.
More details are described in Appendix \ref{sec:auto_annotate}.

Given the skeleton $S=\{s_1, s_2, ..., s_q\}$, 
the pointer network is trained to maximize the conditional log-likelihood:
\begin{align}
\mathcal{L}_1 &=  -\sum_{t=1}^{q+1}{\rm log}\ P_{copy}(s_t|s_{0:t-1},T),
\end{align}
where the special tokens $s_0=\langle BOS\rangle$ and $s_{q+1}=\langle EOS \rangle$ denote the beginning and end of the target skeleton.

\subsection{Stage 2: Surface Realization}
At the surface realization stage, we use
the same encoder as in the skeleton construction stage. 
The decoder is an edit-based Transformer decoder~\citep{gu2019levenshtein} that generates text via insertion and deletion operations.
Different from the original Transformer decoder which predicts the next token in an left-to-right manner, the edit-based decoder can predict tokens simultaneously and independently. In this setting, we can use the full self-attention without causal masking.

\subsubsection{Model Structure}
To perform the insertion and deletion operations, we remove the softmax layer at the top of the Transformer decoder and add three operation classifiers: \textit{Deletion Classifier}, \textit{Placeholder Classifier} and \textit{Token Classifier}.
We denote the outputs of the Transformer decoder as $(\bm{z}_0, \bm{z}_1, ..., \bm{z}_n)$, details of these three classifiers are as follows:
\begin{enumerate}
\item \textit{Deletion Classifier} which predicts for each token whether they should be “deleted"(1) or “kept"(0):
	\begin{equation}\label{equ:delete}　　
	\pi_\theta^{\rm del}(d|i,Y)={\rm softmax}(\bm{W}_{\rm del}\bm{z}_i)
	\end{equation}
\item \textit{Placeholder Classifier} which predicts the number of placeholders [PLH] to be inserted at each consecutive pair:
	\begin{equation}\label{equ:plh}　　
	\pi_\theta^{\rm plh}(p|i,Y)={\rm softmax}(\bm{W}_{\rm plh}[\bm{z}_i;\bm{z}_{i+1}])
	\end{equation}
\item \textit{Token Classifier} which replaces each [PLH] with an actual token:
	\begin{equation}\label{equ:ins_word}　　
	\pi_\theta^{\rm tok}(t|i,Y)={\rm softmax}(\bm{W}_{\rm tok}\bm{z}_i)
	\end{equation}
\end{enumerate}

During decoding, we use the skeleton predicted from the first stage as the initial input of the decoder. 
We also use the full table information from encoder side to mitigate the impact of incomplete skeleton.
As shown in Figure \ref{fig:framework}, the skeleton will pass through the three classifiers sequentially for several iterations. 
Benefiting from the full self-attention, each operation is allowed to condition on the entire skeleton, and thus reduces the probability of hallucinating unfaithful contents in the final text.

\subsubsection{Training}
Following \citet{gu2019levenshtein}, we adopt imitation learning to train our model and simplify their training procedure.
The iterative process of our model will produce various of \textit{intermediate sequences}. To simulate the iterative process, we need to construct the intermediate sequence and provide an optimal operation $\bm{a}^*$ (either oracle insertion $\bm{p}^*$, $\bm{t}^*$ or oracle deletion $\bm{d}^*$) as the supervision signal during training.
Given an intermediate sequence $Y$, the optimal operation $\bm{a}^*$ is computed as follows:
\begin{equation}\label{equ:expert_policy}　　
\bm{a}^*= \mathop{\arg\min}_{\bm{a}}\mathcal{D}(Y^*,\mathcal{E}(Y,\bm{a}))
\end{equation}Here, $\mathcal{D}$ denotes the Levenshtein distance \cite{Levenshtein1965BinaryCC} between two sequences, and $\mathcal{E}(Y,\bm{a})$ represents the output after performing operation $\bm{a}$ upon $Y$.

To improve the training efficiency, we construct the intermediate sequence via a simple yet effective way.  
Given a source table, skeleton and reference $(T, S, Y^*)$, We first calculate the longest common subsequence $X$ between $S$ and $Y^*$, and then construct the intermediate sequence $Y'$ by applying random deletion on $Y^*$ except the part of $X$.
We use $Y'$ to learn the insertion and deletion operations. 
The learning objective of our model is computed as follows:
\begin{align}\label{equ:nar_loss}　　
\mathcal{L}_{\rm edit} &= \mathcal{L}_{\rm ins} + \lambda \mathcal{L}_{\rm del} \\
\begin{split}
\mathcal{L}_{\rm ins} &= -\sum_{p_i^*\in \bm{p}^*}{\rm log}\ \pi_{\theta}^{\rm plh}(p_i^*|i,T,Y') \\ & \quad\, -
\sum_{t_i^*\in \bm{t}^*}{\rm log}\ \pi_{\theta}^{\rm tok}(t_i^*|i,T,Y'')
\end{split} \\
\mathcal{L}_{\rm del} &= -\sum_{d_i^*\in \bm{d}^*}{\rm log}\ \pi_{\theta}^{\rm del}(d_i^*|i,T,Y''')
\end{align}
where $Y''$ is the output after inserting placeholders $\bm{p}*$ upon $Y'$, $Y'''$ is the output by applying the model’s insertion policy $\pi_{\theta}^{\rm tok}$ to $Y''$.\footnote{We do argmax from Equation \eqref{equ:ins_word} instead of sampling.}
$\lambda$ is the hyper parameter.\footnote{In our experiment, $\lambda=1$ gives a reasonable good result.}
\subsubsection{Inference}
As mentioned above, at the inference stage, we use the generated skeleton as the initial input of the decoder.
The insertion and deletion operations will perform alternately for several iterations. 
We stop the decoding process when the current text does not change, or a maximum number of iterations has been reached.

In order to completely retain the skeleton in the generated text, we follow \citet{susanto2020lexically} to enforce hard constraints through forbidding the deletion operation on tokens in the skeleton.
Specially, we compute a \textit{constraint mask} to indicate the positions of constraint tokens in the sequence and forcefully set the deletion classifier prediction for these positions to “keep”. 
The constraint masks are recomputed after each insertion and deletion operation.

\section{Experiment Setups}
\subsection{Datasets}
We conduct experiments on the \textbf{WikiBio} \cite{lebret2016neural} and \textbf{WikiPerson} \cite{wang2018describing} datasets. 
Both datasets aim to generate a biography from a Wikipedia table, but they have different characteristics. 
Their basic statistics are listed in Table \ref{tab:dataset_stat}.
\begin{table}[htbp]
\small   
	\begin{tabular}{llllc}
		\toprule
		\textbf{Dataset}	&\textbf{Train} &\textbf{Valid} &\textbf{Test} &\textbf{Avg Len} \\
		\midrule
		WikiBio &582,657 &72,831 &72,831 &26.1 \\
		WikiPerson &250,186 &30,487 &29,982 &70.6 \\
		\bottomrule
	\end{tabular}
	\caption{Statistics of WikiBio and WikiPerson datasets. \textbf{Avg Len} means the average target length of the datasets.} \label{tab:dataset_stat}
\end{table}
\paragraph{WikiBio} 
This dataset aims to generate the first sentence of a biography from a table.
It is a particularly noisy dataset which has 62\% of the references containing extra information not entailed by the source table \cite{dhingra2019handling}. 
\paragraph{WikiPerson} 
Different from the WikiBio whose reference only contains one sentence, the reference of WikiPerson contains multiple sentences to cover as many facts encoded in the source table as possible. 
In addition, WikiPerson uses heuristic rules to remove sentences containing entities that do not exist in the Wikipedia table, making it cleaner compared to the WikiBio dataset.

\subsection{Evaluation Metrics}
\paragraph{Automatic Metrics} 
For automatic evaluation, we apply BLEU \cite{papineni2002bleu} as well as PARENT (precision, recall, F1) \cite{dhingra2019handling} to evaluate our method.
Different from BLEU which only compare the outputs with the references, PARENT evaluates the outputs that considers both the references and source tables.
Following \citet{wang2020towards}, we further use their proposed PARENT-T metric to evaluate our method in WikiPerson dataset. PARENT-T is a variant of PARENT which only considers the correlation between the source tables and the outputs.
    
\paragraph{Human Evaluation}
Human ratings on the generated descriptions provide more reliable reflection of the model performance.
Following \citet{liu2019towards}, we conduct comprehensive human evaluation between our model and the baselines. The annotators are asked to evaluate from three perspectives: fluency, coverage (how much table content is recovered) and correctness (how much generated content is faithful to the source table). We hire 5 experienced human annotators with linguistic background. During the evaluation, 100 samples are randomly picked from the WikiPerson dataset. For each sample, an annotator is asked to score the descriptions generated by different models without knowing which model the given description is from. The scores are within the range of $[0,4]$.

\subsection{Implementation Details}
We implement SANA using fairseq \cite{ott2019fairseq}. 
The token vocabulary is limited to the 50K most common tokens in the training dataset.
The dimensions of token embedding, key embedding, position embedding are set to 420, 80, 5 respectively.
All Transformer components used in our methods adopt the base Transformer \cite{vaswani2017attention} setting with $d_{\rm model}=512, d_{\rm hidden}=2048, n_{\rm heads}=8, n_{\rm layers}=6$. All models are trained on 8 NVIDIA V100 Tensor Core GPUs.

For the skeleton construction model, the learning rate linearly warms up to 3e-4 within 4K steps, and then decays with the inverse square root scheduler. Training stops after 15 checkpoints without improvement according to the BLEU score. We set the beam size to 5 during inference.

For the surface realization model, the learning rate linearly warms up to 5e-4 within 10K steps, and then decays with the inverse square root scheduler. Training stops when the training steps reach 300K. We select the best checkpoint according to the validation BLEU.

\subsection{Baselines}
We compare SANA with two types of methods: end-to-end methods and two-stage methods. 

For end-to-end methods, we select the following methods as baselines:
(1) \textbf{DesKnow} \cite{wang2018describing}, a seq2seq model with a table position self-attention to capture the inter-dependencies among related attributes; 
(2) \textbf{PtGen} (Pointer-Generator, \citet{see2017get}), an LSTM-based seq2seq model with attention and copy mechanism; 
(3) \textbf{Struct-Aware} \cite{liu2018table}, a seq2seq model using a dual attention mechanism to consider both key and value information; 
(4) \textbf{OptimTrans} \cite{wang2020towards}, a Transformer based model that incorporates optimal transport matching loss and embedding similarity loss.
(5) \textbf{Conf-PtGen} \cite{tian2019sticking}, a pointer generator with a confidence decoding technique to improve generation faithfulness; 
(6) \textbf{S2S+FA+RL} \cite{liu2019towards}, a seq2seq model with a force attention mechanism and a reinforce learning training procedure; 
(7) \textbf{Bert-to-Bert} \cite{Rothe2019LeveragingPC}, a Transformer encoder-decoder model where the encoder and decoder are both initialized with BERT \cite{devlin2019bert}. 

For two-stage methods, we select the following methods as baselines:
(1) \textbf{Pivot} \cite{ma2019key}, a two stage method that first filter noisy attributes in the table via sequence labeling and then uses the Transformer to generate text based on the filter table; 
(2) \textbf{Content-Plan} \cite{puduppully2019data_content}, a two stage method that first uses a pointer network to select important attributes to form a content-plan and then uses a pointer generator to generate text based on the content-plan.

\section{Results}
\begin{table*}[htbp]
\small
  \centering
  \begin{tabular*}{1.0\textwidth}{@{\extracolsep{\fill}}l|ccc|cc}
	\toprule
		\multirow{2}*{\centering \ \ \textbf{Model}} &\multicolumn{3}{c|}{\textbf{WikiPerson}} &\multicolumn{2}{c}{\textbf{WikiBio}} \\
		&\textbf{BLEU} &\textbf{PARENT(P / R / F1)} &\textbf{PARENT-T(P / R / F1)} &\textbf{BLEU} &\textbf{PARENT(P / R / F1)} \\
		\midrule
		\ \ DesKnow &16.20 &63.92 / 44.83 / 51.03 &41.10 / 84.34 / 54.22 &- &- \quad / \quad - \quad / \quad - \quad \\
        \ \ PtGen &19.32 &61.73 / 44.09 / 49.52 &42.03 / 81.65 / 52.62 &41.07 &77.59 / 42.12 / 52.10 \\
        \ \ Struct-Aware &22.76 &51.18 / 46.34 / 46.47 &35.99 / 83.84 / 48.47 &44.93 &74.18 / 43.50 / 52.33 \\
        \ \ OptimTrans &24.56 &62.86 / 48.83 / 53.06 &43.52 / 85.21 / 56.10 &- &- \quad / \quad - \quad / \quad - \quad \\
        \ \ Conf-PtGen &- &- \quad / \quad - \quad / \quad - \quad &- \quad / \quad - \quad / \quad - \quad &38.10 &\textbf{79.52} / 40.60 / 51.38 \\
        \ \ S2S+FA+RL &- &- \quad / \quad - \quad / \quad - \quad &- \quad / \quad - \quad / \quad - \quad &45.49 &76.10 / 43.66 / 53.08 \\
        \ \ Bert-to-Bert &- &- \quad / \quad - \quad / \quad - \quad &- \quad / \quad - \quad / \quad - \quad &45.62 &77.64 / 43.42 / 53.54 \\
		\midrule
		\ \ SANA &\textbf{25.23} &\textbf{65.69} / \textbf{56.88} / \textbf{59.96} &\textbf{44.88} / \textbf{99.47} / \textbf{61.34} &\textbf{45.78} &76.93 / \textbf{46.01} / \textbf{55.42} \\
		\ \ $-$ \textit{hard constrains}  &24.97 &64.72 / 56.42 / 59.29 &43.75 / 98.97 / 60.17 &45.31 &76.32 / 45.26 / 54.64 \\
		\ \ $-$ \textit{skeleton} &19.55 &61.80 / 44.29 / 50.29 &40.80 / 84.03 / 53.97 &42.58 &74.29 / 41.32 / 50.41 \\
  \bottomrule
  \end{tabular*}
    \caption{Comparison with end-to-end methods. \textbf{P, R, F1} represent precision, recall and F1 score, respectively. ``\textit{$-$ hard constrains}" means removing the restriction of forbidding the deletion operation on tokens in the skeleton, ``\textit{$-$ skeleton}" means removing the skeleton construction stage.} \label{tab:ex_end2end}
\end{table*}

\begin{table*}[htbp]
\small
  \centering
  \begin{tabular*}{1.0\textwidth}{@{\extracolsep{\fill}}l|ccc|cc}
	\toprule
		\multirow{2}*{\centering \ \ \textbf{Model}} &\multicolumn{3}{c|}{\textbf{WikiPerson}} &\multicolumn{2}{c}{\textbf{WikiBio}} \\
		&\textbf{BLEU} &\textbf{PARENT(P / R / F1)} &\textbf{PARENT-T(P / R / F1)} &\textbf{BLEU} &\textbf{PARENT(P / R / F1)} \\
		\midrule
		\ \ Pivot &24.71 &62.24 / 50.02 / 52.99 &41.78 / 89.68 / 56.35 &44.39 &76.35 / 41.90 / 51.85  \\
		\ \ + \textit{Oracle} &25.08 &62.34 / 50.63 / 53.47 &42.08 / 89.71 / 56.59 &45.38 &75.98 / 42.57 / 52.45 \\
		\midrule
		\ \ Content-Plan &25.07 &58.56 / 53.86 / 54.52 &38.63 / 91.18 / 54.01 &43.21 &74.69 / 43.53 / 52.71 \\
		\ \ + \textit{Oracle} &28.50 &59.31 / 56.02 / 55.96 &39.64 / 91.62 / 55.07 &50.57 &76.32 / 47.33 / 56.45 \\
		\midrule
        \ \ {SANA} &25.23 &65.69 / 56.88 / 59.96 &44.88 / 99.47 / 61.34 &45.78 &76.93 / 46.01 / 55.42 \\
		\ \ + \textit{Oracle}  &\textbf{30.29} &\textbf{69.27} / \textbf{67.89} / \textbf{68.28} &\textbf{45.13} / \textbf{99.79} / \textbf{61.54} &\textbf{54.51} &\textbf{80.03} / \textbf{51.02} / \textbf{61.01} \\
  \bottomrule
  \end{tabular*}
    \caption{Comparison with two-stage methods. \textbf{P, R, F1} represent precision, recall and F1, respectively. ``\textit{+ Oracle}" means using oracle information (i.e., oracle skeleton or content-plan) as input. } \label{tab:ex_two}
\end{table*}

\subsection{Comparison with End-to-End Methods}

We first compare SANA with end-to-end methods, Table \ref{tab:ex_end2end} shows the experimental results. From Table \ref{tab:ex_end2end}, we can outline the following statements: (1) For WikiPerson dataset, SANA outperforms existing end-to-end methods in all of the automatic evaluation metrics, indicating high quality of the generated texts. Specially, we obtain a near-optimal PARENT-T recall of 99.47, which shows that our model has the ability to cover all the contents of the table. (2) For the noisy WikiBio dataset, SANA outperforms previous state-of-the-art models in almost all of the automatic evaluation scores except the PARENT precision, which confirms the robustness of our method. 
Although Conf-PtGen achieves the highest PARENT precision, its PARENT recall is significantly lower than any other method. Different from Conf-PGen, SANA achieves the highest recall while maintaining good precision. (3) It is necessary to prohibit deleting tokens in the skeleton. After removing this restriction (\textit{$-$ hard constrains}), our method has different degrees of decline in various automatic metrics. (4) SANA performs poorly after removing the skeleton construction stage (\textit{$-$ skeleton}). This shows that the edit-based non-autoregressive model is difficult to directly apply to table-to-text generation tasks. The skeleton is very important for the edit-based model, which can significantly reduce the impact of the multi-modality problem. 
Combining both autoregressive and non-autoregressive generations, SANA achieves state-of-the-art performance.


\subsection{Comparison with Two-Stage Methods}
We further compare SANA with the two-stage methods.
As shown in Table \ref{tab:ex_two}, there is an obvious margin between SANA and the two baselines, which shows that SANA can more effectively model the two-stage process. 
In order to prove that SANA can make use of the information provided by the first stage, we use the gold standard (i.e., the oracle skeleton or content-plan extracted from heuristics methods) as the input of the models used in the second stage. 
With this setup, SANA has made significant improvements in multiple automatic metrics while other methods have limited improvements. 
Specially, the improvements of Pivot are limited because its gold standard does not model the order of the attributes. 
Although the first stage of Content-Plan is similar to SANA, its PARENT scores (either precision, recall and F1) has not been obvious improved, especially on WikiPerson dataset. This shows that the edit-based decoder of SANA can make use of the oracle skeleton to produce high quality descriptions.

\begin{table}[t]
\setlength\tabcolsep{4pt}
\small   
	\begin{tabular}{lccc}
		\toprule
		\textbf{Model}	&\textbf{Fluency} $\uparrow$ &\textbf{Coverage} $\uparrow$ &\textbf{Correctness} $\uparrow$ \\
		\midrule
		Pivot & 3.40 & 3.58 & 2.89  \\
		Content-Plan & 3.39 & 3.70 & 2.98  \\
		Struct-Aware & 3.31 & 3.60 & 2.94  \\
		DesKnow & 3.45 & 3.42 & 3.07  \\
		SANA & \textbf{3.46} & \textbf{3.72} & \textbf{3.11}  \\
		\bottomrule
	\end{tabular}
	\caption{Human evaluation on WikiPerson for SANA and baselines. The scores (higher is better) are based on fluency, coverage and correctness, respectively.} \label{tab:human_eval}
\end{table}

\begin{table*}[t]
\begin{minipage}[t]{\columnwidth}
	\linespread{1.3}\selectfont
	\small
	\centering
	\begin{tabular} {p{435pt}}
		\toprule
		\textbf{(1) Name\_ID}: William Edward Ayrton 
		\textbf{(2) Place\_of\_burial}: Brompton Cemetery 
		\textbf{(3) Place\_of\_birth}: London \\ 
		\textbf{(4) Educated\_at}: University College London 
		\textbf{(5) Date\_of\_birth}: 14 September 1847
		\textbf{(6) Occupation}: Physicist \\
		\textbf{(7) Date\_of\_death}: 8 November 1908  
		\textbf{(8) Aware\_received}: Fellow of the Royal Society 
		\textbf{(9) Child}: Barbara Ayarton-Gould \\
		\bottomrule
	\end{tabular}
\end{minipage}
\\[5pt]
\begin{minipage}[t]{\columnwidth}
	\linespread{1.1}\selectfont
	\small
	\centering
	\begin{tabular} {p{40pt}|p{60pt}|p{310pt}}
		\toprule
		Model &Miss &Output \\
		\hline
        DesKnow &\textbf{(9)}    &William Edward Ayrton Fellow of the Royal Society ( 14 September 1847 – 8 November 1908 ) was a British Physicist . Brompton Cemetery he was born in London {\color{red}the son of Sir Thomas and his wife Mary ( nee Fleming )} . he was educated at {\color{red}University College School} and University College London . \\
		\hline
		Struct-Aware &\textbf{(1) (2) (9)} &William Edward {\color{red}Keeler} Fellow of the Royal Society (14 September 1847 – 8 November 1908) was a British Physicist and Physicist . he was elected a Fellow of the Royal Society {\color{red}in 1889} and was a member of {\color{red}the Royal Society of London} and {\color{red}the Royal Society of London} and {\color{red}the Royal Society of London} . he was educated at the University College London and at the University College London where he was {\color{red}a pupil of the chemist William}. \\
		\hline
		OptimTrans &\textbf{None} &William Edward Ayrton Fellow of the Royal Society (14 September 1847 – 8 November 1908) was an English Physicist . William was born in London and educated at University College London. he is buried in Brompton Cemetery London . he was elected a Fellow of the Royal Society {\color{red}in 1901}. he was the father of Barbara Ayrton-Gould . \\
		\hline
		SANA &\textbf{None} &William Edward Ayrton Fellow of the Royal Society ( 14 September 1847 – 8 November 1908 ) was an English Physicist . he is buried in Brompton Cemetery London . he studied physics at University College London . Ayrton was born in London . he was the father of Barbara Ayrton-Gould. \\
		\bottomrule
	\end{tabular}
\end{minipage}
\caption{Example outputs from different methods. The {\color{red}red} text stands for the hallucinated content in each generated description. Given the table, all the models \textbf{except SANA} generate unfaithful content to varying degrees. Meanwhile, both DesKnow and Struct-Aware miss some table facts, while SANA recovers them all.} \label{tab:case study}
\end{table*}

\subsection{Human Evaluation}

We report the human evaluation result on the WikiPerson dataset in Table~\ref{tab:human_eval}. 
From the demonstrated results, it can be found that SANA outperforms the other end-to-end or two-stage models on all the human evaluation metrics. 
This is consistent with our model's performance in the automatic evaluation. 
In the evaluation of fluency, though the models except for Struct-Aware reach similar performances, SANA performs the best, which demonstrates that its generation has fewer grammatical and semantic mistakes. 
In the evaluation coverage, SANA outperforms the Content-Plan model and defeats the other models by a large margin. 
This result is consistent with our proposal that SANA can cover sufficient information in the source table, and it can ensure the informativeness of generation. 
As to correctness, the advantage of SANA over the other models indicates that our model generates more faithful content and suffers less from the hallucination problem. It should be noted that although Content-Plan and DesKnow are on par with SANA on coverage and correctness respectively, they fail to perform well on both metrics in contrast with SANA. 
This indicates that our model generates both informative and faithful content.

\subsection{Case Study}
Table \ref{tab:case study} shows the descriptions generated by different methods from the test set of WikiPerson.\footnote{For fair comparison, we use the generation examples of baselines provided by \citet{wang2020towards}} 
DesKnow and Struct-Aware miss some attributes and hallucinate unfaithful contents (marked in red). 
Although OptimTrans achieves better coverage, it hallucinates the unfaithful content ``in 1901" not entailed by the table.
Compared to these methods, our method can cover all the attributes in the table and does not introduce any unfaithful contents. 
In addition, the generation length of SANA is shorter than Struct-Aware and OptimTrans, which shows that SANA can use more concise text to cover the facts of the table.
These results indicate our method is capable of generating faithful and informative text.
We put more generation examples in Appendix \ref{sec:more_examples}.

\section{Conclusion}
In this paper, we focus on faithful and informative table-to-text generation. 
To this end, we propose a novel skeleton-based method that combines both autoregressive and non-autoregressive generations. 
The method divides table-to-text generation into skeleton construction and surface realization stages. 
The separated stages helps model better learn the correlation between the source table and reference.
In the surface realization stage, we further introduce an edit-based non-autoregressive model to make full use of the skeleton.
We conduct experiments on the WikiBio and WikiPerson datasets. Both automatic and human evaluations demonstrate the effectiveness of our method, especially on faithfulness and coverage. 


\section*{Acknowledgements}
We thank Tianyu Liu for his suggestions on this research and his providing of inference results of the baseline models. We also thank Yunli Wang for the insightful discussion.

\bibliographystyle{acl_natbib}
\bibliography{anthology,acl2021}






\appendix

\section{Automatic Skeleton Annotation}
\label{sec:auto_annotate}
Algorithm \ref{alg:skeleton_annotation} describes the automatic skeleton annotation process. Given a table and its corresponding description, we first collect tokens appearing in both the table and description except the stop words, then these tokens are sorted based on their positions in the description in ascending order. In this way, we can obtain a sequence composed of the selected tokens. We regard this sequence as a skeleton.

        


\section{More Generation examples}
\label{sec:more_examples}
We further provide a case study, using another two examples (including a very challenging example which needs to recover a large number of facts), to show the effectiveness of our method SANA. In the following pages, we show the example outputs in Table~\ref{tab:case_study_2} and \ref{tab:case_study_3}. In these examples, the SANA model shows much better capability of generating informative and faithful descriptions compared with the baselines.

\begin{algorithm}[H]
    \caption{Automatic Skeleton Annotation}
    \label{alg:skeleton_annotation}
    \begin{algorithmic}[1]
        
        \REQUIRE A stop word set $W$, a parallel dataset $D=\{(T_1,Y_1^*), (T_2, Y_2^*), ..., (T_{|D|},Y_{|D|}^*)\}$;
        \ENSURE A skeleton list $S$;

        \STATE Initial the skeleton list $S=[]$
        \FOR{$(T_i,Y_i^*)\in D$}
        \STATE $T_i = ((k_1,v_1),(k_2,v_2),...,(k_m,v_m))$
        \STATE $V_i = (v_1,v_2,...,v_m)$
        \STATE $Y_i^* = (y_1^*,y_2^*,...,y_n^*)$
        \STATE Initialize the skeleton list $S_i=[]$
        \FOR{$y_j^*\in Y_i^*$}
        \IF {$y_j^*\in V_i$ and $y_j^*\notin W$}
        \STATE Append token $y_j^*$ to the skeleton list $S_i$
        \ENDIF
        \ENDFOR
        \STATE collect the skeleton list $S$ += $S_i$
        \ENDFOR
    \end{algorithmic}
\end{algorithm}

\begin{table*}[t]
\begin{minipage}[t]{\columnwidth}
	\linespread{1.3}\selectfont
	\small
	\centering
	\begin{tabular} {p{435pt}}
		\toprule
		\textbf{(1) Name\_ID}: Aaron Miller
		\textbf{(2) Member\_of\_sports\_team}: Colorado Avalanche
		\textbf{(3) Member\_of\_sports\_team}: Quebec Nordiques 
		\textbf{(4) Date\_of\_birth}: August 11 1971
		\textbf{(5) Place\_of\_birth}: Buffalo, New York
		\textbf{(6) Country\_of\_citizenship}: United States 
		\textbf{(7) Participant\_of}: 2006 Winter Olympics
		\textbf{(8) Participant\_of}: 2002 Winter Olympics
		\textbf{(9) Sport}: Ice hockey \\
		\textbf{(10) Position\_played\_on\_team / speciality}: Cornerback \\
		\bottomrule
	\end{tabular}
\end{minipage}
\\[5pt]
\begin{minipage}[t]{\columnwidth}
	\linespread{1.1}\selectfont
	\small
	\centering
	\begin{tabular} {p{40pt}|p{60pt}|p{310pt}}
		\toprule
		Model &Miss &Output \\
		\hline
        DesKnow &\textbf{7, 8}    &Aaron Miller ( born August 11 1971 ) is an United States former professional Ice hockey Defenceman who played in the National Hockey League ( NHL ) for the Quebec Nordiques and the Colorado Avalanche . he was born in Buffalo, New York and played for the Quebec Nordiques and the {\color{red}Ottawa Senators} . \\
		\hline
		Struct-Aware &\textbf{7, 8} &Aaron Miller (born August 11 1971 in Buffalo, New York New York) is a retired United States professional Ice hockey Defenceman who played in the National Hockey League (NHL) for the Quebec Nordiques Colorado Avalanche Colorado Avalanche Colorado Avalanche Colorado Avalanche and Quebec Nordiques. {\color{red}he was drafted in the 2nd round of overall of the 2002 NHL Entry Draft.} \\
		\hline
		OptimTrans &\textbf{None} &Aaron Miller (born August 11 1971 in Buffalo, New York) is an United States former professional Ice hockey Defenceman who played in the National Hockey League (NHL) for the Quebec Nordiques and Colorado Avalanche . he was a member of the United States men’s national Ice hockey team at the 2002 Winter Olympics and 2006 Winter Olympics. \\
		\hline
		SANA &\textbf{None} & Aaron Miller ( born August 11 1971 in Buffalo, New York ) is a retired United States professional Ice hockey Defenceman . he also played for the Quebec Nordiques and the Colorado Avalanche . Miller was also a member of United States 's ice hockey in the 2002 Winter Olympics and 2006 Winter Olympics . \\
		\bottomrule
	\end{tabular}
\end{minipage}
\caption{Example outputs from different methods. The {\color{red}red} text stands for the hallucinated content in each generated description. Compared with DesKnow and Struct-Aware, SANA recovers all the table facts without generating any unfaithful content.} \label{tab:case_study_2}
\end{table*}

\begin{table*}[t]
\begin{minipage}[t]{\columnwidth}
	\linespread{1.3}\selectfont
	\small
	\centering
	\begin{tabular} {p{435pt}}
		\toprule
		\textbf{(1) Name\_ID}: Émile Mbouh
		\textbf{(2) Member\_of\_sports\_team}: Le Havre AC
		\textbf{(3) Member\_of\_sports\_team}: Perlis FA
		\textbf{(4) Member\_of\_sports\_team}: Sport Benfica e Castelo Branco
		\textbf{(5) Member\_of\_sports\_team}: Qatar SC
		\textbf{(6) Member\_of\_sports\_team}: Vitória S.C. 
		\textbf{(7) Member\_of\_sports\_team}: Tanjong Pagar United FC
		\textbf{(8) Member\_of\_sports\_team}: CS Chênois
		\textbf{(9) Member\_of\_sports\_team}: Cameroon national football team 
		\textbf{(10) Number\_of\_matches\_played}: 46 
		\textbf{(11) Number\_of\_points / goals / set\_scored}: 2
		\textbf{(12) Member\_of\_sports\_team}: Union Douala 
		\textbf{(13) Member\_of\_sports\_team}: Diamant Yaoundé
		\textbf{(14) Member\_of\_sports\_team}: Ettifaq FC
		\textbf{(15) Member\_of\_sports\_team}: Liaoning Whowin F.C.
		\textbf{(16) Member\_of\_sports\_team}: Sabah FA 
		\textbf{(17) Place\_of\_birth}: Douala 
		\textbf{(18) Date\_of\_birth}: 30 May 1966 
		\textbf{(19) Sport}: Association football
		\textbf{(20) Participant\_of}: 1994 FIFA World Cup 
		\textbf{(21) Participant\_of}: 1990 FIFA World Cup
		\textbf{(22) Position\_played\_on\_team / speciality}: Midfielder \\
		\bottomrule
	\end{tabular}
\end{minipage}
\\[5pt]
\begin{minipage}[t]{\columnwidth}
	\linespread{1.1}\selectfont
	\small
	\centering
	\begin{tabular} {p{40pt}|p{60pt}|p{310pt}}
		\toprule
		Model &Miss &Output \\
		\hline
        DesKnow &\textbf{2, 3, 4, 5, 6, 8, 10, 11, 12, 13, 14, 15, 16, 21, 22}    &Émile Mbouh ( born 30 May 1966 ) is a former Cameroon national football team Association football . he was born in Douala and played for the Tanjong Pagar United FC in the 1994 FIFA World Cup . \\
		\hline
		Struct-Aware &\textbf{2, 3, 4, 5, 6, 8, 10, 11, 12, 13, 14, 15, 16, 17, 21} & Émile Mbouh, (born 30 May 1966) is a Cameroonian retired Association football who played as a Midfielder . {\color{red}Le} represented Cameroon national football team at the 1994 FIFA World Cup and 1994 FIFA World Cup . he played for {\color{red}Le FC Sport Yaoundé, United Yaoundé} and Tanjong Pagar United FC\\
		\hline
		OptimTrans &\textbf{2, 3, 5, 6, 8, 12, 13, 14} &Émile Mbouh (born 30 May 1966) is a Cameroonian retired Association football who played as a Midfielder . born in Douala Émile began his career with Sport Benfica e Castelo Branco and Tanjong Pagar United FC . he also represented Cameroon national football team at the 1994 FIFA World Cup and 1990 FIFA World Cup . he also played for Sabah FA and Liaoning Whowin F.C. in the Malaysia Super League . he also played for Tanjong Pagar United FC and Liaoning Whowin F.C. in the Chinese Super League. \\
		\hline
		SANA &\textbf{3, 12, 13, 21} & Émile Mbouh ( born 30 May 1966 ) is a retired Cameroonian Association football who played as a Midfielder . born in Douala Mbouh played club football in France for Sport Benfica e Castelo Branco Le Havre AC CS Chênois Vitória S.C. Tanjong Pagar United FC Qatar SC Ettifaq FC Tanjong Pagar United FC Qatar SC Sabah FA and Liaoning Whowin F.C. . Mbouh played for the Cameroon national football team ( 46 caps and scoring 2 goals ) and two games at the 1994 FIFA World Cup . \\
		\bottomrule
	\end{tabular}
\end{minipage}
\caption{Example outputs from different methods. The {\color{red}red} text stands for the hallucinated content in each generated description. This table contains a large number of facts to recover, which makes the case very challenging. In contrast with the other models, SANA misses much fewer facts and does not produce unfaithful content.} \label{tab:case_study_3}
\end{table*}

\end{document}